# Alleviating Class Imbalance in Semi-supervised Multi-organ Segmentation via Balanced Subclass Regularization

Zhenghao Feng, Lu Wen, Binyu Yan, Jiaqi Cui, Yan Wang*, *Member, IEEE*

*Abstract*—Semi-supervised learning (SSL) has shown notable potential in relieving the heavy demand of dense prediction tasks on large-scale well-annotated datasets, especially for the challenging multi-organ segmentation (MoS). However, the prevailing class-imbalance problem in MoS, caused by the substantial variations in organ size, exacerbates the learning difficulty of the SSL network. To alleviate this issue, we present a two-phase semi-supervised network (BSR-Net) with balanced subclass regularization for MoS. Concretely, in Phase I, we introduce a class-balanced subclass generation strategy based on balanced clustering to effectively generate multiple balanced subclasses from original biased ones according to their pixel proportions. Then, in Phase II, we design an auxiliary subclass segmentation (SCS) task within the multi-task framework of the main MoS task. The SCS task contributes a balanced subclass regularization to the main MoS task and transfers unbiased knowledge to the MoS network, thus alleviating the influence of the class-imbalance problem. Extensive experiments conducted on two publicly available datasets, i.e., the MICCAI FLARE 2022 dataset and the WORD dataset, verify the superior performance of our method compared with other methods.

## I. INTRODUCTION

Multi-organ segmentation (MoS) [1, 2, 3, 4], which aims to simultaneously assign an accurate class label to each pixel of multiple organs inside the radiology images, is an imperative task in computer-assisted diagnosis [5, 6, 7, 8]. Recently, deep learning (DL)-based segmentation methods have reached promising results with the fully supervised training on massive labeled data [9, 10, 11, 12]. However, gathering ample annotated data for such data-driven methods is unrealistic due to the expensive time and labor costs.

To reduce the reliance on annotations, semi-supervised learning (SSL) enhances the segmentation performance by utilizing both the limited labeled data and abundant unlabeled data [13, 14, 15]. For instance, based on the popular SSL architecture, i.e., mean teacher [12], [15] employs dual-level contrastive learning strategies to explore the pixel-wise and organ-wise correlations. [16] utilizes an attention mechanism to force the model to focus more on the regions of interest (ROIs)

This work is supported by National Natural Science Foundation of China (NSFC 62371325, 62071314), Sichuan Science and Technology Program 2023YFG0263, 2023YFG0025, 2023YFG0101. Zhenghao Feng and Lu Wen contribute equally to this work. Corresponding author: Yan Wang. Zhenghao Feng, Lu Wen, Binyu Yan, Jiaqi Cui, and Yan Wang are with the School of Computer Science, Sichuan University, China. (e-mail: fzh_scu@163.com; wenlu0416@163.com; yanby@scu.edu.cn; jiaqicui2001@gmail.com; wangyanscu@hotmail.com).

inside the nasopharyngeal carcinoma. Besides, [37] builds a semi-supervised segmentation model with variance-reduced estimation to promote the performance with extremely limited labels. [17] uses an extra regression task to learn richer feature to refine the segmentation results. However, most works mainly focus on single-organ segmentation in a semi-supervised manner, limiting practical applicability in clinical settings. Thus, semi-supervised MoS (SSMoS) naturally comes to sight.

One crucial problem in SSMoS is class imbalance arising from substantial differences in the size of organs. The model trained on the class-imbalance data may bias to the larger organs, leading to lower accuracy for the smaller ones [18]. Currently, several class-rebalance strategies have been explored, i.e., re-weighting [19, 38], re-sampling [20], and meta-learning [21, 39]. [19] presented a class adaptive Dice loss to balance the penalties to different ROIs based on their pixel proportions. [22] designed a cascade of decision trees to largely decrease the number of large targets. [20] explored the impact of different sampling methods, e.g., oversampling, and undersampling, on the final accuracies. However, these strategies have two main limitations. First, they mainly focus on fully supervised settings where labeled data are required to correct the biased predictions, and are thus not applicable to unlabeled data in SSMoS. Second, re-weighting or resampling methods lack further generation or utilization of the balanced data, limiting further performance enhancements. So, it is essential to develop an effective solution to relieve the class-imbalance problem in SSMoS task.

In this paper, to alleviate the above issues, we propose a two-phase semi-supervised network (BSR-Net) that utilizes a balanced subclass regularization to learn unbiased knowledge for the MoS task. Specifically, in phase I, to priorly mine the latent balanced information, we use a class-balanced subclass generation strategy to produce multiple balanced subclasses from original biased classes. Subsequently, in Phase II, we construct an auxiliary subclass segmentation (SCS) task within the multi-task framework to provide an additional class-balanced regularization of the main MoS network, thus gradually transferring unbiased knowledge from the SCS network to the MoS network.

Overall, the paper makes the contributions as follows: (1) We introduce a novel two-phase semi-supervised network, called BSR-Net, to effectively utilize the unlabeled data for the challenging SSMoS task. (2) We present a balanced subclass regularization accompanied with an auxiliary SCS task to incorporate the class-unbiased knowledge into the main MoS task in a multi-task framework, thus relieving the class-imbalance problem. (3) Extensive experiments verify the superior segmentation performance of the proposed method compared to those of other state-of-the-art methods both quantitatively and qualitatively.

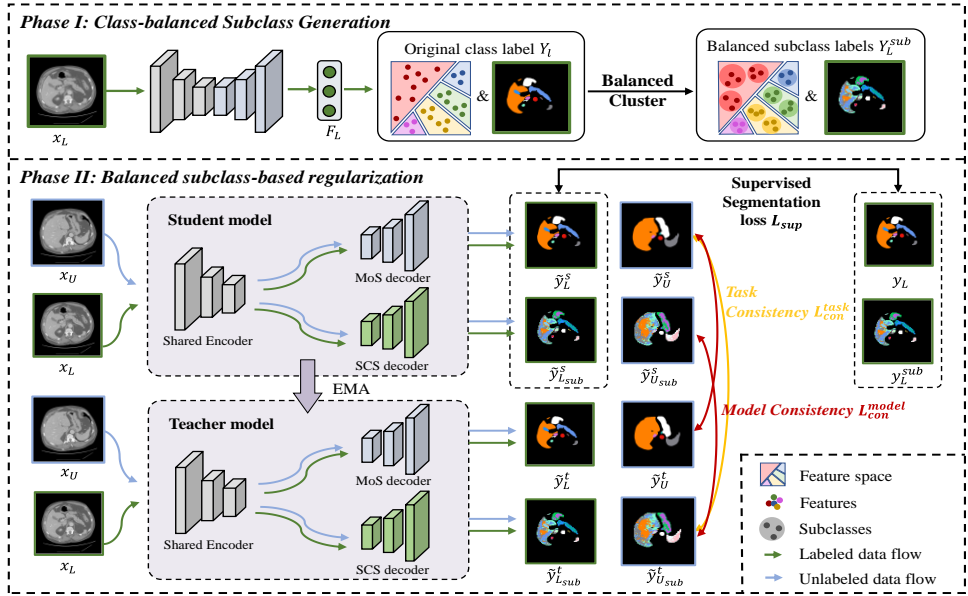

**Fig. 1.** Illustration of the proposed BSR-Net.

## II. METHODOLOGY

The architecture of the proposed two-phase BSR-Net is depicted in Fig 1. Concretely, in Phase I, we utilize the labeled data to pre-trained the backbone and then use the well-trained backbone to produce the balanced subclasses through a balanced clustering, thus mining the latent unbiased knowledge inside the original labels. In Phase II, we build a semi-supervised network based on the mean teacher [12] framework. To employ such balanced information, we follow the idea of multi-task learning [6, 23, 24] and construct an auxiliary subclass segmentation (SCS) task besides the MoS task. The two tasks are incorporated with a shared encoder and two task-specific decoders. Finally, the output of the SCS network with abundant balanced knowledge provides a balanced subclass regularization to the main MoS network and enforces it to focus more on small targets, thus enhancing the overall accuracy.

In our problem setting, the labeled set is represented as $D_L = \{(x_L^i, y_L^i)\}_{i=1}^{N}$ where the $x_L^i \in R^{H \times W}$ represents the radiation image of height $H$ and width $W$, and $y_L^i \in \{0, 1 \dots K\}^{H \times W}$ is the segmentation labels with $K$ total organ substructures (0 means background) to be segmented. The unlabeled set is defined as $D_U = \{x_U^i\}_{i=N+1}^{N+M}$ where $N \ll M$. Network details are stated in the following subsections.

### A. Phase I: Class-balanced Subclass Generation

Considering the class-imbalance problem caused by the large size differences among different organs, we design a class-balanced subclass generation strategy to separate the original classes into several class-balanced subclasses with almost equal pixel numbers. Concretely, we adopt U-net [25] as the backbone and train it with the labeled set $D_L$ with a supervised segmentation loss, thus enabling it with the fundamental ability of feature extraction. To perform pixel clustering and generate balanced data, we omit the output layer in the pre-trained backbone and map the labeled image $x_L$ into pixel-level semantic features $F_L = \{f_i\}_{i \in [1,p]}$, where $p$ represents the total pixel number ($C$ for channel) and $f_i$ denotes the feature vector of $i$-th pixel. Next, we conduct a clustering operation on the feature vectors where the vectors belonging to the same class are aggregated together to form a cluster, which is then considered as a subclass. Notably, balanced clustering [34], unlike the traditional clustering methods, e.g., k-means clustering [26], adjusts the pixel number in each cluster based on the pixel proportions of the original classes. Thus, the larger targets are divided into more subclasses while the smaller ones gain fewer subclasses, resulting in multiple subclasses with nearly equal numbers of pixels. Once all the original classes have been re-divided, a new balanced subclass label $y_{L_{sub}} \in \{0, 1 \dots K_{sub}\}^{H \times W}$ is obtained, where $K_{sub}$ is the total number of subclasses. Subsequently, the class-balanced labeled dataset $D_{L_{sub}} = \{(x_L^i, y_{L_{sub}}^i)\}_{i=1}^{N}$ is utilized to perform an additional regularization in Phase II.

### B. Phase II: Balanced Subclass Regularization

In Phase II, we design a SSMoS network with a balanced subclass-based regularization. Inspired by the notable performance of the mean teacher which contains a student and a teacher model [12], we avail it as the backbone of Phase II. To utilize the unbiased knowledge in the class-balanced data $D_{L_{sub}}$, following the idea of multi-task learning, we design a main MoS task and an auxiliary SCS task where the two tasks are incorporated with a shared encoder and two task-specific decoders. Then, the output of the SCS task can provide class-balanced regularization to the main MoS task, thus transferring the unbiased knowledge from the SCS network to the MoS one.

**Student Model.** Following Phase I, we employ the U-net [25] as the backbone for both the main MoS task and auxiliary SCS task. Notably, the encoder is shared by the two tasks while the parameters in the two task-specific decoders are different to fit different tasks. In this manner, the encoder is also enforced to capture the crucial features associated with small structures during the optimization process. Therefore, fed with a labeled image $x_L$ (unlabeled image $x_U$), the two subnetworks produce the MoS prediction $\tilde{y}_L^s$ ($\tilde{y}_U^s$) and SCS prediction $\tilde{y}_{L_{sub}}^s$ ($\tilde{y}_{U_{sub}}^s$):

$$\tilde{y}_L^s = f_{mos}(x_L; \theta_{mos}, \varepsilon), \tilde{y}_{L_{sub}}^s = f_{scs}(x_L; \theta_{scs}, \varepsilon), \quad (1)$$
$$\tilde{y}_U^s = f_{mos}(x_U; \theta_{mos}, \varepsilon), \tilde{y}_{U_{sub}}^s = f_{scs}(x_U; \theta_{scs}, \varepsilon), \quad (2)$$

where $f_{mos}$ and $f_{scs}$ denote the MoS and SCS network with corresponding parameters $\theta_{mos}$ and $\theta_{scs}$, respectively, and $\varepsilon$ represents the data perturbation in the student model.

**Teacher Model.** The teacher model follows the same architecture as the student model and updates its parameters, i.e., $\theta'_{mos}$ and $\theta'_{scs}$, by exponential moving average (EMA) [12]. Similarly, inputted with an image $x_L$ ($x_U$), the teacher model also outputs the MoS prediction $\tilde{y}_L^t$ ($\tilde{y}_U^t$) and SCS prediction $\tilde{y}_{L_{sub}}^t$ ($\tilde{y}_{U_{sub}}^t$) with the following formulation:

$$\tilde{y}_L^t = f_{mos}(x_L; \theta'_{mos}, \varepsilon'), \tilde{y}_{L_{sub}}^t = f_{scs}(x_L; \theta'_{scs}, \varepsilon'), \quad (3)$$
$$\tilde{y}_U^t = f_{mos}(x_U; \theta'_{mos}, \varepsilon'), \tilde{y}_{U_{sub}}^t = f_{scs}(x_U; \theta'_{scs}, \varepsilon'), \quad (4)$$

where $\varepsilon'$ is the data perturbation in the teacher model. Then, the predictions made by the teacher model can serve as the additional supervisions for those of the student model.

**Balanced Subclass Regularization.** As mentioned in Section II.A, the subclass labels are priorly subdivided from the original ones, so the main MoS and auxiliary SCS task theoretically maintain the same semantic information. Based on this, we propose a task consistency loss, i.e., $L_{con}^{task}$, to perform the balanced subclass regularization between these two tasks. Specifically, we map the predicted subclass predictions, i.e., $\tilde{y}_{U_{sub}}^t$, to the original class, i.e., $y'_U$, and supervise the MoS predictions, which is expressed as follows:

$$y'_U = map(\tilde{y}_{U_{sub}}^t), \quad (5)$$
$$L_{con}^{task} = L_{ce}(y'_U, \tilde{y}_U^s) + L_{dice}(y'_U, \tilde{y}_U^s), \quad (6)$$

where the $map(\cdot)$ represents the mapping function. In this regularization way, we embed the unbiased knowledge in the balanced subclass into the main MoS network, thus effectively enhancing the model's attention to the small targets.

*C. Objective Functions*

To constrain the predictions of the student, i.e., $\tilde{y}_L^s$ and $\tilde{y}_{L_{sub}}^s$, via labeled data, we impose the following supervised loss:

$$L_{sup} = L_{seg}(y_L, \tilde{y}_L^s) + \alpha L_{seg}(y_{L_{sub}}, \tilde{y}_{L_{sub}}^s), \quad (7)$$

where $L_{seg}$ also equally incorporates two classical pixel-wise losses, i.e., cross-entropy (CE) loss $L_{ce}$ and Dice loss $L_{dice}$.

Following the design of mean teacher [12], we introduce the model consistency loss $L_{con}^{model}$ to force the prediction of an unlabeled input $x_U$ from the student to keep similar to that from the teacher, which is formulated as follows:

$$L_{con}^{model} = L_{mse}(\tilde{y}_U^s, \tilde{y}_U^t) + L_{mse}(\tilde{y}_{U_{sub}}^s, \tilde{y}_{U_{sub}}^t), \quad (8)$$

where $L_{mse}$ means a mean-square error (MSE) loss.

Therefore, the whole loss function can be written as the weighted sum of $L_{sup}$, $L_{con}^{model}$, and task consistency loss $L_{con}^{task}$:

$$L_{total} = L_{sup} + \beta_1 L_{con}^{model} + \beta_2 L_{con}^{task}, \quad (9)$$

where $\beta_1$ and $\beta_2$ are the weighted terms.

## III. EXPERIMENTS AND RESULTS

*A. Datasets and Evaluation*

**MICCAI Flare 2022 Dataset** is a subset of the abdomen computed tomography (CT) image segmentation Flare challenge [27] to alleviate the domain shifts among multiple centers [28, 29]. It contains 135 CT volumes. There are 13 organs needed to be segmented: Liver (LV), Right kidney (RK), Spleen (SP), Pancreas (PA), Aorta (AO), Inferior Vena Cava (IVC), Right Adrenal Gland (RAG), Left Adrenal Gland (LAG), Gallbladder (GB), Esophagus (ES), Stomach (ST), Duodenum (DU), and Left kidney (LK). We randomly select 100/10/25 samples as training/validation/ testing set.

**WORD Dataset** is a large-scale Whole abdominal Organ Dataset [36] with 150 CT volumes. Besides 9 shared organs with Flare dataset, i.e., LV, LK, RK, SP, PA, ST, GB, DU, and ES, there are 7 specific organs needed to be segmented: colon (CO), intestine (IN), adrenal (Adr), rectum (RE), bladder (BL), left head of the femur (LH), and right head of the femur (RH). We follow the official partitions which use 100/20/30 samples as training/ validation/testing set.

In the training set, we further divide the labeled set and the unlabeled set as $n/m$ to simulate the semi-supervised setting, where $n$ and $m$ are the numbers of labeled and unlabeled samples. We employ two commonly used metrics, i.e., Dice coefficient and Jaccard Index (JI), to quantitatively measure the overlapping between the prediction and the ground truth.

*B. Implementation Details*

We conduct experiments with the PyTorch framework and trained on a single NVIDIA GeForce RTX 3090 GPU with a total memory of 24GB. SGD optimizer is employed to train the whole model for 20000 iterations with a learning rate of 1e-2 and batch size of 16. $\alpha$ in Eq. (7) is empirically set as 0.1. Following [20], $\beta_1$ is set as 0.1. $\beta_2$ is set to 0 in the first 5000 iterations for the instability of subclass segmentation. For the remaining 15000 iterations, its value is chosen with hyper-parameter selection experiments on the validation set of WORD dataset. Specifically, when $\beta_2$ is set as 0.01, 0.05, 0.1, 0.5, and 1, we respectively gain 70.41%, 71.7%, 72.86%, 73.93% and 73.33% mean Dice. So, we set $\beta_2$ as 0.5. Moreover, the teacher model is chosen as the final prediction model for its better stability and generalization.

*C. Comparative Experiments*

To verify the performance of our proposed method in SSMoS, we compare it with six state-of-the-art (SOTA) methods, i.e., U-net (2015) [25], mean teacher (MT, 2017) [12], uncertainty aware mean teacher (UAMT, 2019) [30], interpolation consistency training (ICT, 2022) [31], uncertainty rectified pyramid consistency (URPC, 2022) [32], and evidential inference learning (EVIL, 2024) [33]. Notably, we only report the results of the largest three organs and the smallest three organs for page limitation. As seen in Table I, the proposed gains the best overall performance for all the data partitions. Concretely, when only 5 labeled data is available, our method surpasses the second-best EVIL by 4.60% mean Dice and 5.97% JI, and achieves 73.68% for ES, 64.38% for LAG, 66.56% for RAG in terms of Dice. As the labeled data increases, the proposed method, URPC, and EVIL all perform well on the largest LV with a fewer performance disparities and our method maintains its leading performance for the small organs, thus finally getting the best mean accuracy. When n=10 and 15, our method performs relatively bad on Es which may results from the inaccurate and inconsistent annotations in the manual process. The visualizations are shown in Fig.2 where our method can accurately segment the targets with the least fault segmentation. Concretely, for n=5, the proposed gains the best segmentation performance on small targets marked by red arrows while other methods have remarkable wrong predictions.

**TABLE I.** QUANTITATIVE COMPARISON WITH SIX SOTA METHODS IN TERMS OF DICE AND JI WHEN N=5, 10, AND 15 RESPECTIVELY. THE BEST RESULTS ARE IN **BOLD** WHILE THE SECOND-BEST ONES ARE <u>UNDERLINED</u>.

| n/m | Methods | LV Dice | LV JI | ST Dice | ST JI | SP Dice | SP JI | ES Dice | ES JI | LAG Dice | LAG JI | RAG Dice | RAG JI | Mean Dice | Mean JI |
|---|---|---|---|---|---|---|---|---|---|---|---|---|---|---|---|
| 5/95 | U-net [18] | 88.95 | 80.12 | 67.86 | 51.53 | 70.83 | 55.34 | 62.48 | 45.57 | 58.32 | 41.37 | 58.83 | 41.99 | 63.50 | 48.06 |
| | MT [5] | 93.56 | 87.92 | 71.89 | 56.36 | 80.46 | 67.78 | 68.82 | 52.76 | 52.59 | 35.96 | 55.71 | 39.24 | 67.51 | 53.78 |
| | UAMT [21] | 91.51 | 84.38 | 66.15 | 49.59 | 85.76 | 75.30 | 67.21 | 50.71 | 58.79 | 41.88 | 62.97 | 46.49 | 68.07 | 53.87 |
| | ICT [22] | 92.21 | 85.56 | <u>74.51</u> | <u>59.52</u> | <u>89.35</u> | <u>80.83</u> | 68.80 | 52.55 | 60.82 | 44.45 | 63.00 | 46.25 | 70.02 | 56.07 |
| | URPC [23] | 92.39 | 85.89 | 63.24 | 46.79 | 86.32 | 76.41 | <u>69.49</u> | <u>53.44</u> | <u>62.16</u> | <u>45.18</u> | 61.89 | 45.21 | 69.93 | 55.90 |
| | EVIL [24] | <u>93.98</u> | <u>88.65</u> | 70.03 | 54.14 | 88.89 | 80.02 | 69.26 | 53.14 | 59.54 | 42.61 | <u>67.82</u> | <u>51.53</u> | <u>72.21</u> | <u>58.71</u> |
| | Proposed | **94.51** | **90.88** | **74.51** | **61.72** | **93.01** | **88.44** | **73.68** | **56.28** | **64.38** | **47.52** | **66.56** | **50.14** | **76.81** | **64.68** |
| 10/90 | U-net [18] | 94.41 | 89.43 | 80.00 | 66.77 | 92.77 | 86.59 | 72.11 | 56.43 | 70.63 | 54.65 | 70.01 | 54.06 | 77.58 | 65.67 |
| | MT [5] | 95.92 | 92.16 | 80.17 | 66.97 | 95.70 | 91.76 | 73.95 | 58.82 | 62.59 | 45.67 | 68.53 | 52.70 | 78.60 | 66.99 |
| | UAMT [21] | 95.37 | 91.14 | 80.39 | 67.27 | 94.25 | 89.13 | 71.25 | 55.44 | <u>71.48</u> | <u>55.75</u> | 74.15 | 59.68 | 79.67 | 68.31 |
| | ICT [22] | 96.43 | 93.11 | <u>84.61</u> | <u>73.41</u> | 94.92 | 90.34 | <u>76.81</u> | <u>62.41</u> | 69.63 | 53.48 | <u>74.74</u> | <u>60.08</u> | 80.68 | 69.67 |
| | URPC [23] | **96.86** | **93.90** | 82.92 | 70.89 | 95.35 | 91.13 | 75.93 | 61.33 | 69.92 | 53.87 | **76.23** | **61.93** | 81.04 | 70.02 |
| | EVIL [24] | <u>96.52</u> | <u>93.28</u> | 81.45 | 69.12 | <u>96.47</u> | <u>93.21</u> | **79.58** | **66.13** | 69.72 | 53.93 | 71.08 | 55.84 | <u>81.66</u> | <u>70.96</u> |
| | Proposed | 96.39 | 93.04 | **85.95** | **75.41** | **96.72** | **93.65** | 76.54 | 62.09 | **71.74** | **56.06** | 71.26 | 56.88 | **83.06** | **72.85** |
| 15/85 | U-net [18] | 97.20 | 94.56 | 87.56 | 77.89 | 96.46 | 93.17 | 78.57 | 64.73 | 71.17 | 55.35 | 77.59 | 63.57 | 84.35 | 74.61 |
| | MT [5] | 96.34 | 92.95 | 81.79 | 69.22 | 96.50 | 93.25 | 80.49 | 67.42 | 76.43 | 61.91 | 76.66 | 62.44 | 84.72 | 74.86 |
| | UAMT [21] | 95.85 | 92.03 | 84.33 | 72.97 | 96.30 | 92.86 | 76.72 | 62.26 | 74.37 | 59.37 | 75.56 | 60.95 | 83.99 | 73.89 |
| | ICT [22] | 97.35 | 94.83 | 86.40 | 76.08 | 96.70 | 93.62 | 79.29 | 65.75 | 75.94 | 61.26 | <u>79.50</u> | <u>66.13</u> | 85.27 | 75.59 |
| | URPC [23] | **97.55** | **95.21** | <u>89.99</u> | <u>81.83</u> | <u>96.81</u> | <u>93.82</u> | <u>81.73</u> | <u>69.13</u> | 74.75 | 59.74 | 76.62 | 62.48 | 86.04 | 76.89 |
| | EVIL [24] | <u>97.49</u> | <u>95.11</u> | 89.51 | 81.05 | 90.38 | 82.47 | **82.68** | **70.51** | <u>78.51</u> | <u>64.70</u> | 79.15 | 65.75 | <u>86.36</u> | <u>77.26</u> |
| | Proposed | 96.91 | 94.00 | **90.38** | **82.47** | **96.90** | **94.00** | 80.98 | 68.07 | **78.63** | **64.81** | **81.02** | **68.28** | **87.07** | **78.16** |

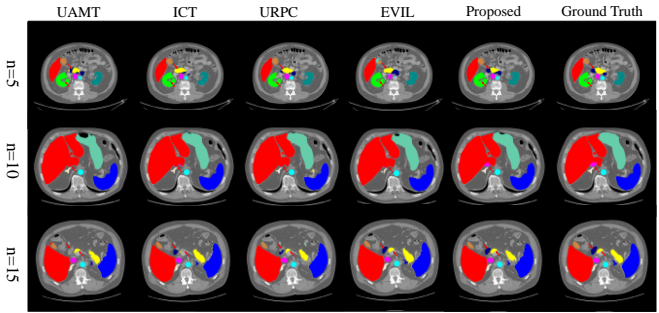

**Fig. 2.** Visualization comparisons with SOTA models.

**TABLE II.** QUANTITATIVE RESULTS OF SOTA METHODS ON WORD DATASET WHEN N=5 IN TERMS OF DICE.

| Methods | LV | SP | LK | BL | LH | RH | Mean |
|---|---|---|---|---|---|---|---|
| URPC[23] | 91.13 | 76.98 | 83.90 | 82.94 | 85.06 | 89.67 | 67.46 |
| EVIL[24] | 92.16 | 76.69 | 82.81 | 83.14 | 88.77 | 88.07 | 68.75 |
| Proposed | **94.17** | **84.69** | **85.55** | **90.05** | **88.90** | **90.83** | **72.79** |

To further validate the model generalizability, Table II presents the results of URPC, EVIL, and our proposed on the WORD dataset. As seen, for larger targets, with the uncertain estimation, both URPC and EVIL perform well on large organs. However, compared to EVIL, our method still gains 2.01%, 8.00%, and 2.74% higher Dice. Besides, for the three smaller organs, our method also performs even better than other two methods. Especially, our method obtains 7.11% and 6.91% promotions than URPC and EVIL in terms of Bla.

These above-mention experimental results have verified the effectiveness of our method on addressing the class-imbalance SSMoS both quantitatively and qualitatively.

*D. Ablation study*

We conduct several ablation experiments to investigate the effectiveness of the key components in the proposed method. The experimental arrangements can be summarized as: (A) labeled data only, (B) mean teacher +traditional model consistency loss (MT + $L_{con}^{model}$), (C) MT + $L_{con}^{model}$ + balanced SCS task, (D) MT + $L_{con}^{model}$ + imbalanced SCS task + task consistency loss ($L_{con}^{task}$), and (E) MT + $L_{con}^{model}$ + balanced SCS task + task consistency loss ($L_{con}^{task}$) (proposed). The quantitative results are displayed in Table III where the gradually increased performance is gained by progressively adding the key components. Concretely, compared (C) to (B), the balanced SCS task enhance the Dice by 13.61% and 6.77% in terms of LAG and RAG, respectively. Then, further adding the task consistency loss, (E) improves the overall Dice from 71.89% to 76.81%, indicating its effectiveness in promoting the main MoS task. Furthermore, to further validate the effect on the balanced subclasses, we additionally design model (D) which utilizes k-means clustering [26] to generate subclasses for the SCS task. As seen, compared with (D), (E) obtains 8.61% and 4.13% accuracy enhancements in terms of ES and SP, demonstrating the positive impact of balanced subclasses on the segmentation of small organs.

**TABLE III** ABLATION STUDY OF OUR METHOD IN TERMS OF DICE.

| Methods | LV | ST | SP | ES | LAG | RAG | Mean |
|---|---|---|---|---|---|---|---|
| (A) | 88.95 | 67.86 | 70.83 | 62.48 | 58.33 | 58.83 | 63.50 |
| (B) | 93.56 | 71.89 | 80.46 | 68.82 | 52.59 | 55.71 | 67.51 |
| (C) | 92.53 | 70.76 | 85.59 | 68.96 | **66.20** | 62.48 | 71.87 |
| (D) | 92.47 | 71.74 | 88.88 | 65.07 | 60.56 | 64.81 | 72.06 |
| (E) | **94.51** | **74.51** | **93.01** | **73.68** | 64.38 | **66.56** | **76.81** |

## VI. CONCLUSION

In this paper, we present the BSR-Net, a semi-supervised network with balanced subclass regularization to tackle the class-imbalance issue in SSMoS. By generating the class-balanced subclasses with a balanced cluster in Phase I and introducing a subclass segmentation auxiliary task to provide balanced subclass regularization to the main MoS task in Phase II, we are able to effectively transfer the unbiased knowledge to the MoS model and enhance the prediction accuracy of small organs. Extensive experiments on two abdominal MoS datasets have verified the superiority of our method.